# Exploring the importance of context and embeddings in neural NER models for task-oriented dialogue systems

**Pratik Jayarao**
Haptik Inc, Mumbai
pratik.jayarao@haptik.co

**Chirag Jain**
Haptik Inc, Mumbai
chirag.jain@haptik.co

**Aman Srivastava**
Haptik Inc, Mumbai
aman.srivastava@haptik.co

## Abstract

Named Entity Recognition (NER), a classic sequence labelling task, is an essential component of natural language understanding (NLU) systems in task-oriented dialog systems for slot filling. For well over a decade, different methods from lookup using gazetteers and domain ontology, classifiers over hand crafted features to end-to-end systems involving neural network architectures have been evaluated mostly in language-independent non-conversational settings. In this paper, we evaluate a modified version of the recent state of the art neural architecture in a conversational setting where messages are often short and noisy. We perform an array of experiments with different combinations of including the previous utterance in the dialogue as a source of additional features and using word and character level embeddings trained on a larger external corpus. All methods are evaluated on a combined dataset formed from two public English task-oriented conversational datasets belonging to travel and restaurant domains respectively. For additional evaluation, we also repeat some of our experiments after adding automatically translated and transliterated (from translated) versions to the English only dataset.

## 1 Introduction

Named Entity Recognition (NER) is a challenging and vital task for slot-filling in task-oriented dialogue models. We define a task-oriented dialogue system where a user and the system take turns exchanging information till some concluding action is performed related to user's query. Any task oriented bot must figure out the correct intent and fill slots for requested task interactively until all required slots required for the related action are filled. For most domains and languages, a very small amount of supervised data is available. Generalizing from such small amount of data can be challenging as for some entities are open ended in what values they can assume like NAME while for some other entities like CITY or LOCATION it is very likely some values will appear very rarely while training. As a result, a lot of hand-crafted features and domain-specific knowledge resources and gazetteers are used for solving this task.

Unfortunately, collecting such resources is time-consuming for each new domain and language making a cold start even harder. Most work for NER has been benchmarked on the popular CoNLL2003 dataset (Sang and Meulder, 2003), OntoNotes 5.0 and few other datasets. Only very recently high quality medium to large sized task-oriented dialogue English datasets like DSTC2 (Henderson et al., 2014), Frames (Asri et al., 2017), etc. have been made available to focus on the challenge of dialogue state tracking. Some of these datasets also happen to have slots tagged hence making them suitable to benchmark NER systems. We combined two such datasets - DSTC2 (Restaurant table booking system) and Frames (Airline ticket booking system) to evaluate our approaches in multi-domain multi-entity setting. We observed that in conversational datasets, the systems usually yield longer informative messages and users often tend to provide information in multiple short messages. For example,

**System:** What city are you flying to?
**User:** Paris

In such cases, the immediately previous system utterance provides important context regarding the domain and slots to predict. Users also tend to make spelling mistakes which can create problem for models using words as unit features.

In the past few years, end-to-end neural architectures (Huang et. al., 2015; Lample et al., 2016;

Ma and Hovy, 2016) with a CRF layer have shown promising results for NER. Our work is inspired from these models where we too use an end-to-end BI-LSTM-CRF based tagger network but also include an additional LSTM based context encoder that encodes the system occurrence immediately before the user's query which we use to initialize the tagger network's initial state. We observe that this additional context improves F1 score for all settings we test. Word embedding features (Mikolov et al., 2013; Pennington et al., 2014) obtained using unsupervised training on a large corpora have also shown to improve results (Huang et. al., 2015; Lample et al., 2016; Ma and Hovy, 2016).

Following that we too explore different initializations of word embeddings in our experiments. We also compute additional word representation from character level representations using Convolutional Neural Networks (CNNs) and combine them with pre-trained word embeddings. To validate that our models can work in language-independent settings, we also present results after adding automatically translated (to Hindi) and transliterated (to English back from Hindi) versions of the datasets and find that results follow the same trends as results on the English only dataset.

## 2 Related Work

NER is viewed as a sequential prediction problem. Most work related to this task if often benchmarked on CoNLL2003 and OntoNotes datasets. Early work includes typical models like HMM (Rabiner, 1989), CRF (Lafferty et al., 2001), and sequential application of Perceptron or Winnow (Collins, 2002). (Ratinov and Roth, 2009) highlight design challenges involved in NER and achieve an F1 of 90.80 on CoNLL2003 using averaged perceptron model trained on non-local features, gazetteers extracted from Wikipedia and brown clusters extracted from large corpora. (Lin and Wu, 2009) get a better score using linear chain CRF with L2 regularization without any gazetteers but instead by obtaining phrase features from clustering a massive dataset of search query logs. (Passos et al., 2014) match their performance by using a linear chain CRF on hand-crafted features and phrase vectors trained using a modified semi-supervised skip-gram architecture. (Luo et al., 2015) jointly model NER and entity linking tasks. They include hand-engineered features like spelling features, lexical clusters, shallow parsing features as well as stemming and large external knowledge bases.

A shift in trend towards more neural based approaches can be traced back to (Collobert et al., 2011b) proposed an effective deep neural model with a CRF layer on top, that requires almost no feature engineering and learns important features from word embeddings trained on large corpora. The model achieved near state of the art results on several natural language tasks including NER. (Santos and Guimaraes, 2015) modify this architecture to incorporate character level features computed using CNNs and report better F1 scores on both CoNLL2003 and OntoNotes. Coming to architectures that involve recurrent neural networks, (Huang et. al., 2015) test LSTM, BI-LSTM, LSTM-CRF and BI-LSTM-CRF networks with word embeddings and hand-crafted spelling and context features for several sequence tagging tasks. (Lample et al., 2016) also use BI-LSTM-CRF models but don't rely on any hand-crafted features and external resources. They use pre-trained word embeddings and character level embeddings computed from another BI-LSTM network as primary features. (Ma and Hovy, 2016) use similar architecture but compute character level embeddings using CNN layer instead of BI-LSTMs. (Chiu and Nichols, 2015) also work with very similar BI-LSTM-CNN-CRF architecture with some extra character level features like character type, capitalization, and lexicon features. Finally, very recent work from (Peters et. al., 2018) show improvements on NER task by computing better contextualized word level embeddings features.

Our work is closely related to (Ma and Hovy, 2016) in terms of the neural architecture such that we too work in an end-to-end setting and include character level features using a CNN. Although we experimented with different recurrent cell types for our models due to space limitations we present only BI-LSTM models. Our work mostly focuses on using conversational context as a source of additional features, determining optimal word embeddings representations for the task.

## 3 Dataset

We primarily use data from two publicly available dialogue datasets Maluuba Frames (Asri et al., 2017) and Dialog State Tracking Challenge 2 (DSTC2) (Henderson et al., 2014).

To evaluate the system for English we constructed first dataset DSTC-FRAMES-EN by combining the two datasets to get a total of 13599 user-system utterance pairs and 7 entities ('or_city', 'dst_city', 'budget', 'date', 'area', 'food', 'price_range'). We split this data into 12000 train samples and 1599 test samples with a total of 412 unique entities (with IOB-prefixes).

We also check if our models can work with more than one language simultaneously. For this we translate DSTC-FRAMES-EN to Hindi using Google Translate (Wu et al., 2016) and further transliterate this translated version to Latin script using Polyglot transliteration (Chen and Skiena, 2016). We combine all three to form an extended dataset DSTC-FRAMES-ENHI which contains a total of 37785 samples, 7 entities with 1106 unique entities values (with IOB-prefixes). We split this combined dataset into 34000 training samples and 3785 test samples.

## 4 Model Architecture

### 4.1 Input Layer

We use two kinds of input representations - word embeddings and character embeddings concatenated with word representations.

**Word Embeddings:** Word Embeddings have a significant role in the increasing the performance of various neural inspired models as they exploit the syntactic and semantic understanding of a word. We conducted experiments with four different (frozen) embeddings w.r.t. dimension size, demographics of training data and size of training data.

**Skip Gram Negative Sampling (SG300)**: We trained 300 dimensional word embeddings separately using the SGNS (Mikolov et al., 2013) model. We restricted the training data to training set only for each experiment.

**Glove:** We used the publicly available Glove embeddings[1] (Pennington et al., 2014) trained on Wikipedia 2014 corpus with dimension sizes 50 (**G50W**) and 300 (**G300W**) and another 300 dimensional embeddings (**G300C**) trained on a significantly larger Common Crawl dataset.

**Character Embeddings (CHAR):** Character level features can be useful to handle rare words and spelling errors which are usually OOV for word embedding models. To use this character-level knowledge, we employ a convolutional neural network (CNN) with 30 filters of fixed window of size 3. We perform max pooling on output of convolution operations to generate 100 dimensional embeddings for each word. This character-level representation is then concatenated with the corresponding word embedding and fed into the network. Such character level embeddings have been shown to have potential to replace hand crafted character features. (Chiu and Nichols, 2015)

[1] https://nlp.stanford.edu/projects/glove/

### 4.2 Context Encoder (CE)

The context encoder we implemented is a unidirectional LSTM network. At every time step tokens from the latest system utterance are fed as inputs to the context encoder. The encoder updates its internal state thus transforming the system utterance to a rich fixed sized representation which is then fed to the tagger's forward hidden state. The context encoder enables the system to scale across various domain and language settings by maintaining the immediate history.

### 4.3 Tagger

The Tagger network is responsible for performing the NER on user utterances.

**Recurrent Neural Network (RNN):** RNN (Elman, 1990; Ubeyli and ¨Ubeyli, 2012) consists of a hidden state that depending on the previous hidden state and current input continually updates itself at every time step. The output is then predicted on the basis of the new hidden state.

**Long Short Term Memory (LSTM):** LSTMs (Hochreiter and Schmidhuber, 1997) a modification to RNNs introduce additional gated mechanisms to manage the vanishing/exploding gradient problems faced by RNN.

**Bidirectional LSTM (BI-LSTM):** In NER, at a given time step we have access to both past and future inputs. This gives us an opportunity to implement a BI-LSTM architecture.

### 4.4 Conditional Random Fields (CRF)

We add a linear chain CRF (Lafferty et al., 2001) network on top of the output states (concatenated forward and backward states at each time step) yielded by the tagger network to form BI-LSTM-CRF model. This layer considers dependencies across output labels to compute the log likelihood of IOB sequence tags using Viterbi decoding algorithm.

| Model | SGNS300 | G50W | G300W | G300C |
|---|---|---|---|---|
| BI-LSTM | 86.928 | 88.138 | 89.388 | 90.057 |
| BI-LSTM-CE | 89.130 | 90.163 | 90.910 | 91.224 |
| BI-LSTM-CHAR | 87.465 | 89.089 | 89.442 | 90.551 |
| BI-LSTM-CHAR-CE | 89.412 | 91.087 | 91.342 | 91.880 |
| BI-LSTM-CRF | 87.782 | 89.529 | 89.871 | 90.627 |
| BI-LSTM-CRF-CE | 89.696 | 91.122 | 91.455 | 92.133 |
| BI-LSTM-CHAR-CRF | 88.276 | 89.628 | 90.971 | 91.079 |
| BI-LSTM-CHAR-CRF-CE | **90.036** | **91.705** | **92.042** | **92.864** |

Table 1: Macro Averaged F1 scores on the DSTC-FRAMES-EN dataset

## 5 Experiments

We evaluated the performance of 3 major recurrent neural cell (RNN, GRU, LSTM) types on DSTC-FRAMES-EN by constructing unidirectional recurrent networks using these cells. The LSTM, GRU (Cho et al., 2014) based networks showed significant improvement over RNN architecture. The LSTM architecture displayed a marginal increase in performance in comparison to the GRU network. We thus conducted our further experiments by using the LSTM cell.

All BI-LSTM networks presented are stacked two layers deep with each cell containing 64 hidden units. We train our models with Adam (Kingma and Ba, 2014) optimizer for up to 30 epochs and use early stopping to avoid overfitting.

Since there are no pre-trained Glove embeddings for Hindi and transliterated Hindi, we restricted our set of experiments on DSTC-FRAMES-ENHI to only SGNS embeddings we trained separately

**Choice of Architecture:** As shown in Table 1 and Table 2, the character level embeddings helped increase the performance of the network by leveraging character level features and handling OOV tokens. With an addition of slightly more parameters the CRF layer boosted the system's performance. The networks which included the context encoder showed significant improvements in comparison to their non-context encoder counterparts.

**Choice of word embeddings:** From Table 1, we can see that the G50W displayed better performance than SG300 owing to a larger corpus of training data. The G300W being trained on the same training corpus as G50W exhibited improved results on account of larger dimension size. The best results were displayed by G300C which as trained on a larger and better suited style of data for our conversational settings.

| Model | SGNS300 |
|---|---|
| BI-LSTM | 84.867 |
| BI-LSTM-CE | 86.242 |
| BI-LSTM-CHAR | 85.119 |
| BI-LSTM-CHAR-CE | 86.433 |
| BI-LSTM-CRF | 85.342 |
| BI-LSTM-CRF-CE | 86.790 |
| BI-LSTM-CHAR-CRF | 85.643 |
| BI-LSTM-CHAR-CRF-CE | **87.934** |

Table 2: Macro Averaged F1 scores on the DSTC-FRAMES-ENHI dataset

## 6 Conclusion and Future Work

We presented results for NER from variants of the popular BI-LSTM architecture in a task-oriented conversational setting. Adding a CRF layer boosts performance at slightly extra computational cost. Our results show that context in form of system utterance just before the user query potentially has important information about domain and slots and including it further boosts performance. Although we only included just one utterance from conversational history, including more conversation history can be helpful but can also be challenging as it might put pressure on the context encoder to ignore already detected slots. Nevertheless, it remains to be explored for future work. We also find that NER models also benefit from large pre-trained word representations and character level representations.